%% file: main.tex
\documentclass{article}
\usepackage{microtype}
\usepackage{graphicx}
\usepackage{subfigure}
\usepackage{booktabs}
\usepackage{hyperref}

\usepackage[accepted]{icml2025} 
\usepackage{amsmath}
\usepackage{amssymb}
\usepackage{mathtools}
\usepackage{amsthm}
\usepackage[capitalize,noabbrev]{cleveref}
\usepackage[textsize=tiny]{todonotes}
\usepackage{caption}
\usepackage{comment}
\usepackage{enumitem}

\usepackage{tikz}
\usepackage{adjustbox}

\usepackage{etoolbox}
\AtBeginEnvironment{tabular}{\scriptsize} 

\definecolor{blue}{HTML}{4878D0}
\definecolor{violet}{HTML}{956CB4}
\definecolor{green}{HTML}{6ACC64}
\definecolor{orange}{HTML}{EE854A}
\definecolor{red}{HTML}{D65F5F}
\newcommand{\green}[1]{\textcolor{green}{#1}}
\newcommand{\blue}[1]{\textcolor{blue}{#1}}

\newcommand{\orange}[1]{\textcolor{orange}{#1}}

\icmltitlerunning{Transformers Pretrained on Procedural Data Contain Modular Structures for Algorithmic Reasoning}

\begin{document}

\onecolumn
\icmltitle{Transformers Pretrained on Procedural Data Contain\\Modular Structures for Algorithmic Reasoning}

\icmlsetsymbol{equal}{*}

\begin{icmlauthorlist}
\icmlauthor{Zachary Shinnick}{aiml}
\icmlauthor{Liangze Jiang}{epfl,idiap}
\icmlauthor{Hemanth Saratchandran}{aiml}
\icmlauthor{Anton van den Hengel}{aiml}
\icmlauthor{Damien Teney}{idiap}
\end{icmlauthorlist}

\icmlaffiliation{aiml}{Australian Institute for Machine Learning (AIML), University of Adelaide, Australia}
\icmlaffiliation{epfl}{École Polytechnique Fédérale de Lausanne (EPFL), Switzerland}
\icmlaffiliation{idiap}{Idiap Research Institute, Switzerlad}

\icmlcorrespondingauthor{Zachary Shinnick}{zachary.shinnick@adelaide.edu.au}

\vskip 0.3in


\printAffiliationsAndNotice{}  

\input{abstract}

\input{introduction}

\input{methodology}

\input{results}

\input{discussion}


\clearpage
\bibliography{references}
\bibliographystyle{icml2025}

\newpage
\appendix

\input{relatedWork}

\input{appendix}

\end{document}

%% file: abstract.tex
\begin{abstract}
\vspace{6pt}
\textbf{Context.}
Pretraining on large, semantically rich datasets is key for developing language models.
Surprisingly, recent studies have shown that even synthetic data, generated procedurally through simple semantic-free algorithms, can yield some of the same benefits as natural language pretraining. It is unclear \textit{what} specific capabilities such simple synthetic data instils in a model, \textit{where} these capabilities reside in the architecture, and \textit{how} they manifest within its weights. 

\textbf{Findings.} 
In this short paper, we
identify several beneficial forms of procedural data,
together with specific algorithmic reasoning skills that improve in small transformers.
Our core finding is that different procedural rules instil
\textit{distinct but complementary inductive structures} in the model.
With extensive ablations and partial-transfer experiments,
we discover that these structures reside in different parts of the model.
Attention layers often carry the most transferable information, but some pretraining rules
impart useful structure to MLP blocks instead.
Most interestingly, the structures induced by multiple rules can be composed to jointly reinforce multiple capabilities.

\textbf{Implications.}
These results suggest an exciting possibility of disentangling the acquisition of
knowledge from reasoning in language models, 
with the goal of improving their robustness and data efficiency.
\end{abstract}

%% file: introduction.tex
\section{Introduction}

What properties of pretraining data enable language models to acquire both knowledge and reasoning capabilities? While the size and diversity of the data are empirically crucial for learning factual knowledge~\cite{longpre2024pretrainer}, far less is understood about the structural and distributional characteristics required to acquire reasoning abilities.


\textbf{The value of structured data.}~
Empirically, it has been observed that pretraining language models on computer code has proven highly effective, likely due to the inherent compositional and recursive structure present in code \cite{petty2024does}.
Recent work suggests that procedural data, generated from simple algorithm rules%
\setcounter{footnote}{3} 
\footnote{We distinguish \emph{procedural} data, generated from explicit algorithms, from \emph{synthetic} data, from learned models e.g.\ other LLMs.}
can offer similar benefits.
For instance, \citet{hu2025between} demonstrate that formal language data provides greater value per token than natural language when training a 1B-parameter model. Likewise, \citet{zhang2024intelligence} show that data generated by cellular automata can accelerate the learning of abstract reasoning tasks and yield modest improvements in chess move prediction.

\textbf{This paper.}~
We aim to better understand the mechanisms at play when pre-training
transformer-based sequence models
on procedural data (see Figure~\ref{fig:teaser}).
We characterise and locate useful structures created in such pretrained models
that facilitate subsequent fine-tuning and generalisation 
on algorithmic reasoning and language modelling tasks.
The literature contains many results on
the pretraining of language models with data from formal languages
and 
simple algorithms (see Appendix~\ref{app:related-work}), down to a simple identity function \cite{wu2022insights},
i.e.\ simply learning to repeat the input.
It is conceivable that such pretraining improves over a random initialisation simply 
by adjusting the overall magnitude of the weights~\cite{huang2020improving}.
We design experiments to distinguish such trivial effects
from the genuine learning of transferrable mechanisms.

\input{figTeaser}

\textbf{Summary of findings.}~
We bring elements of answers to the following questions.
\begin{enumerate}[leftmargin=*, itemsep=2pt, parsep=0pt, topsep=-5pt]
\item \textbf{What} is gained from procedural data? (Section~\ref{sec:distinct_capabilities})
We pretrain small transformers on diverse forms of procedural data and fine-tune them on a series of diagnostic tasks.
\textit{We find that specific types of procedural data significantly enhance particular capabilities for algorithmic reasoning.}

\item \textbf{Where} does the useful knowledge reside in the pretrained models? (Section~\ref{sec:partial_transfer})
We evaluate the partial transfer of pretrained weights.
\textit{We find that the beneficial effects of pretraining lie in different parts of the architecture}, depending on the task,
with the most pronounced benefits often found in the attention layers.

\item \textbf{How} does the pretraining improve these downstream capabilities? (Section~\ref{sec:structure})
We evaluate various perturbations of the pretrained weights
to understand how they encode useful information.
For some tasks (e.g.\ \textsc{Sorting}), even shuffled weights preserve \textit{some} benefits,
but all tasks undergo a significant performance drop
with any kind of perturbation (e.g.\ Gaussian noise).
This indicates that \textit{pretraining creates non-trivial inductive structures}
in attention and/or MLP weights.
Moreover,
\textit{these structures prove remarkably modular}. 
The attention layers of a pretrained model can be combined with MLPs of another to construct a better initialization that improves multiple algorithmic reasoning capabilities (Section~\ref{sec:modular}).

\end{enumerate}
\vspace{1pt}
We discuss exciting potential applications of these results in Section~\ref{sec:discussion}.
A longer version of this paper is also in preparation.

%% file: figTeaser.tex
\setlength{\fboxsep}{0pt}     
\setlength{\fboxrule}{0.8pt}  

\setlength{\arraycolsep}{0pt}
\renewcommand{\arraystretch}{1.7} 

\begin{figure}[t!]
    \begin{tabular}{ccc} 
      \raisebox{-.5\height}{\includegraphics[height=0.13\linewidth]{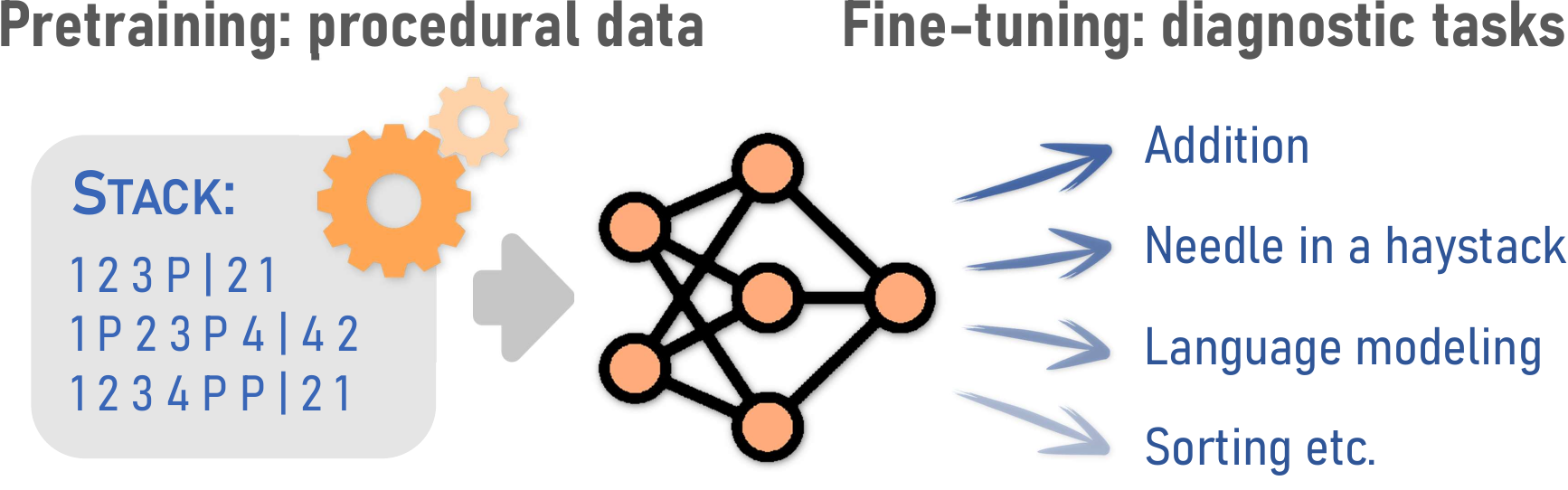}}\hspace{14pt} &
    
      \renewcommand{\arraystretch}{0.9}
      \setlength{\arraycolsep}{0pt}
      \raisebox{-.1\height}{
            \begin{tabular}{ll}
                \toprule
                Pretraining task & Example sequence \\
                \midrule
                \textsc{$k$-Dyck}
                & \texttt{(} \blue{\texttt{[}} \orange{\texttt{\{}} \orange{\texttt{\}}} \blue{\texttt{]}} \texttt{)} \\
                \textsc{$k$-Dyck Shuffle}
                & \texttt{(} \blue{\texttt{[}} \orange{\texttt{\{}} \blue{\texttt{]}} \texttt{)} \orange{\texttt{\}}} \\
                \textsc{Stack} &
                \texttt{1} \texttt{2} \texttt{3} \texttt{P} \texttt{|} \textbf{\texttt{2}} \textbf{\texttt{1}} \\
                \textsc{Identity} &
                \texttt{1} \texttt{2} \texttt{3} \texttt{|} \textbf{\texttt{1}} \textbf{\texttt{2}} \textbf{\texttt{3}} \\
                \textsc{Set} &
                \texttt{1} \texttt{2} \texttt{2} \texttt{|}  \textbf{\texttt{1}} \textbf{\texttt{2}} \\
                \bottomrule
            \end{tabular}
      }
      &
    
      \raisebox{-0\height}{
        \hspace{10pt}
        \begin{tikzpicture}[baseline=(current bounding box.center)]
          \draw[->, thick] (2.3,0.75) -- (2.3,-0.8) node[midway, right=2pt] {Time};
          \node[anchor=west] at (0,0) {\includegraphics[height=0.115\linewidth]{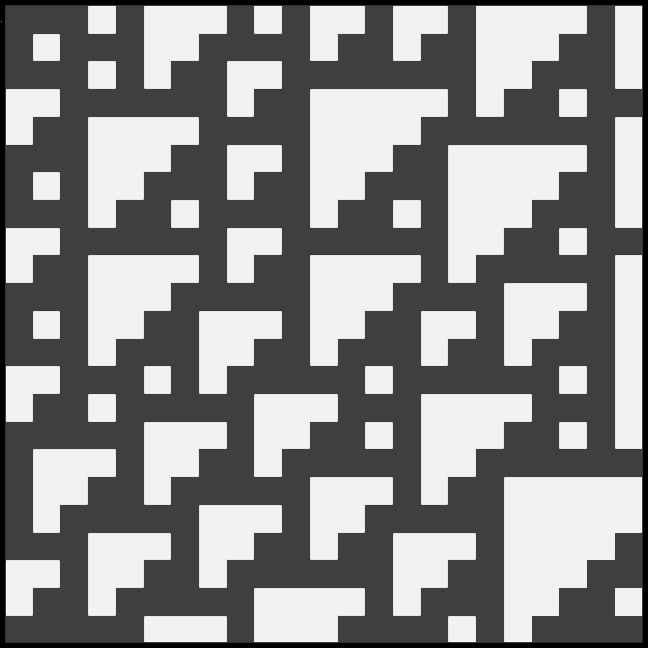}}; 
        \end{tikzpicture}
      }
      \\
      (a)\hspace{17pt} & (b) & \hspace{-10pt}(c)~ECA Rule 110
    \end{tabular}
    \vspace{-7pt}
    \caption{
        (a)~We pretrain small transformers on various forms of procedural data, then fine-tune them on a series of diagnostic tasks.~
        The data is generated from formal languages (b) or simple algorithms such as elementary cellular automata~(c).
        In~\textsc{$k$-Dyck} examples, matching brackets are color-coded.
        For \textsc{Stack}, `\texttt{P}' is the \textit{pop} operation.
        For \textsc{Stack}, \textsc{Identity}, and \textsc{Set},
        `\texttt{|}' acts as a separator between the input and the expected output,
        on which the loss is computed (\textbf{\texttt{bold}} tokens).
    }
    \label{tab:procedural-data-examples}
    \label{fig:teaser}
    \vspace{-10pt}
\end{figure}

\renewcommand{\arraystretch}{1.0} 
\setlength{\arraycolsep}{5pt}

%% file: methodology.tex
\vspace{-1pt}
\section{Methodology}
For each experiment, we train a small transformer on one form of procedural data then fine-tune it on one diagnostic task.

\vspace{-2pt}
\textbf{Procedural pretraining.}~
We pretrain the model for standard next-token prediction
on data generated from formal languages and simple algorithms (Figure~\ref{fig:teaser}a).
The following selection is motivated by prior work on procedural data.
See Appendix~\ref{app:experiment-details} for details.
(1)~\textsc{$k$-Dyck}: a formal language of nested brackets.
(2)~\textsc{$k$-Dyck shuffle}: a variant with matching braces that are not necessarily nested.
(3)~\textsc{Stack}, a simulation of a stack memory; the model must predict the final contents of the memory.
(4)~\textsc{Identity}: the model must repeat the input.
(5)~\textsc{Set}: the model must remove repeated tokens from the input.
(6)~Complex elementary cellular automaton (\textsc{ECA rule 110}): a binary sequence with deterministic Markovian evolution.

\vspace{-2pt}
\textbf{Fine-tuning on diagnostic tasks.}~
Prior work with procedural data focused on language modelling (Appendix~\ref{app:related-work}) while we aim to identify specific improved skills.
We thus fine-tune each model on tasks that span a range of algorithmic reasoning capabilities: memory recall (\textsc{Needle-in-a-haystack}), arithmetic (\textsc{Addition}, \textsc{Reversed addition}, \textsc{Multiplication}), sorting (\textsc{Sorting}), and handling of natural language (\textsc{Language modelling}). See Appendix~\ref{app:downstream_tasks} for details.

\vspace{-2pt}
\textbf{Architecture.}~
Our focus is on the data, hence we use a simple GPT-2-type architecture~\citep{radford2019language}, with 2~layers, 4~attention heads, and a hidden size of 16, or a larger configuration for more challenging tasks (\textsc{Multiplication}, \textsc{Language modeling}).
See Appendix~\ref{app:pretrain_details} and~\ref{sec:downstream-training-details} for details on hyperparameters.

\vspace{-2pt}
\textbf{Weight transfer.}~
We denote the weights of a transformer model as \( \mathcal{T}\!=\!(\mathbf{E}, \mathbf{A}, \mathbf{F}) \)
where \( \mathbf{E} \) corresponds to the embedding and unembedding layers (tokens and positions), \( \mathbf{A} = \{\mathbf{A}^{(1)}, \dots, \mathbf{A}^{(L)}\} \)
to the attention, and
\( \mathbf{F} = \{\mathbf{F}^{(1)}, \dots, \mathbf{F}^{(L)}\} \)
to the MLPs across \( L \) layers.
When the vocabulary of the pretraining and fine-tuning tasks cannot be aligned,
we reset the token embeddings to average pretrained vector \cite{hewitt2021initializing}, 
as an unbiased initialisation before fine-tuning.
The positional embeddings are transferred if the pretrained context length is sufficient,
or randomly initialised otherwise.

%% file: results.tex
\vspace{-1pt}
\section{Different Procedural Tasks Improve Distinct Reasoning Capabilities}
\label{sec:distinct_capabilities}

Prior work on procedural data
demonstrates its value 
for general language modelling~\cite{hu2025between}.
However it is not clear what exact skills improve in this setting.
In this section, 
we show that different forms of procedural data
improve specific capabilities for algorithmic reasoning.
See Appendix~\ref{app:experiment-details}--\ref{app:additional-results}
for additional details and results.


\textbf{Setup.}~
We first pretrain a model
\( \mathcal{T}_{\text{pre}}\!=\!(\mathbf{E}_{\text{pre}}, \mathbf{A}_{\text{pre}}, \mathbf{F}_{\text{pre}}) \)
on procedural data.
We then transfer all weights from \( \mathcal{T}_{\text{pre}} \) into a new model
\( \mathcal{T}_{\text{full}}\!=\!(\mathbf{E}_{\text{pre}}, \mathbf{A}_{\text{pre}}, \mathbf{F}_{\text{pre}}) \),
referred to as ``full-model'' transfer.
\( \mathcal{T}_{\text{full}} \)
is then fine-tuned on a downstream diagnostic task.
We compare its performance to a randomly initialised baseline
\( \mathcal{T}_{\text{rand}}\!=\!(\mathbf{E}_{\text{rand}}, \mathbf{A}_{\text{rand}}, \mathbf{F}_{\text{rand}}) \)
that undergoes the same fine-tuning. 
For \textsc{Language modelling}, we increase the hidden size to 64.
For \textsc{Multiplication}, we use the larger \texttt{gpt2-mini} architecture%
\footnote{\url{https://huggingface.co/erwanf/gpt2-mini}}
to enable a comparison with a 
model pretrained on OpenWebText.

\begin{figure}[h!]
    \vspace{-8pt}
    \centering
    \includegraphics[width=1.0\linewidth]{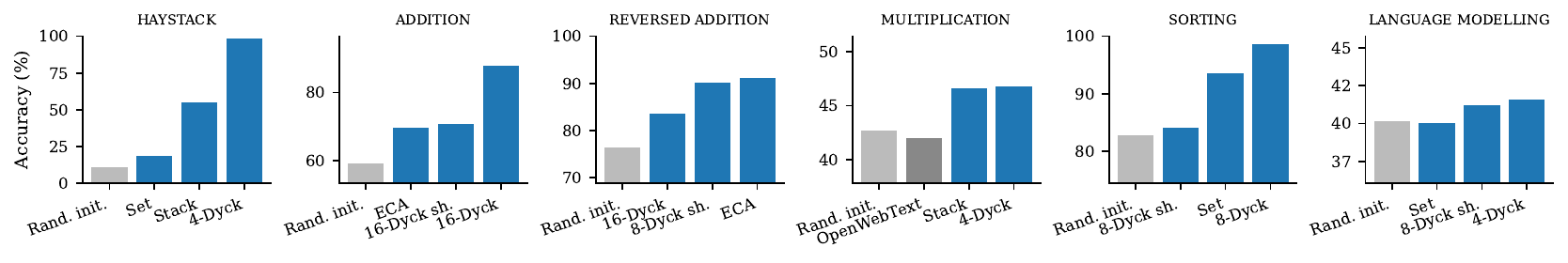}
    \vspace{-26pt}
    \caption{
    Models pretrained on carefully-chosen procedural data (blue) significantly
    improve specific downstream diagnostic tasks,
    compared to a random initialisation (light gray) or pretraining on natural language (dark gray, shown for \textsc{Multiplication}).
    See Appendix~\ref{app:additional-results} for full results,
    including cases where procedural pretraining brings no benefit.
    }
    \label{first-figure}
    \vspace{-7pt}
\end{figure}


\textbf{Results.}~
Figure~\ref{first-figure} shows that procedurally pretrained models can
largely outperform randomly initialized ones for every downstream task,
confirming that \textbf{procedural pretraining data can instil meaningful ``soft inductive biases''}
in transformers.
Furthermore, we observe that the different forms of procedural data
yield different improvements across the downstream tasks.
This indicates that \textbf{each procedural task imparts distinct inductive biases}
that are each relevant to different downstream tasks.
For example, for \textsc{Haystack}, pretraining on \textsc{$k$-Dyck} can inject an effective memory recall capability (in order to match the nested brackets),
which cannot be provided by \textsc{Set}, \textsc{$k$-Dyck shuffle} or \textsc{ECA} (see also Table~\ref{tab:full_res}).
Notably, for \textsc{Multiplication}, procedural pretraining outperforms
pretraining on OpenWebText.


\vspace{-2pt}
\section{Characterising the Effects of Procedural Pretraining}

\vspace{-1pt}
\subsection{The Benefits from Pretraining Often Reside in Specific Architectural Components}
\label{sec:partial_transfer}
\vspace{-1pt}

Procedural pretraining creates useful ``soft inductive biases'' in a model, but where do they reside?
This section shows that
different forms of procedural data
act on different parts of the architecture.

\textbf{Setup for selective transfer.}~
Given a procedurally-pretrained model
\( \mathcal{T}_{\text{pre}}\!=\!(\mathbf{E}_{\text{pre}}, \mathbf{A}_{\text{pre}}, \mathbf{F}_{\text{pre}}) \)
from Section~\ref{sec:distinct_capabilities},
we initialise models for fine-tuning with selected components:
attention-only transfer
\( \mathcal{T}_{\text{attn}}\!=\!(\mathbf{E}_{\text{rand}}, \mathbf{A}_{\text{pre}}, \mathbf{F}_{\text{rand}}) \),
MLP-only transfer
\( \mathcal{T}_{\text{mlp}}\!=\!(\mathbf{E}_{\text{rand}}, \mathbf{A}_{\text{rand}}, \mathbf{F}_{\text{pre}}) \),
or full-model transfer
\( \mathcal{T}_{\text{full}}\!=\!(\mathbf{E}_{\text{pre}}, \mathbf{A}_{\text{pre}}, \mathbf{F}_{\text{pre}}) \).
In all cases, the \emph{entire} model is fine-tuned.

\begin{figure}[h!]
\begin{center}
    \vspace{-9pt}
    \includegraphics[width=1.0\linewidth]{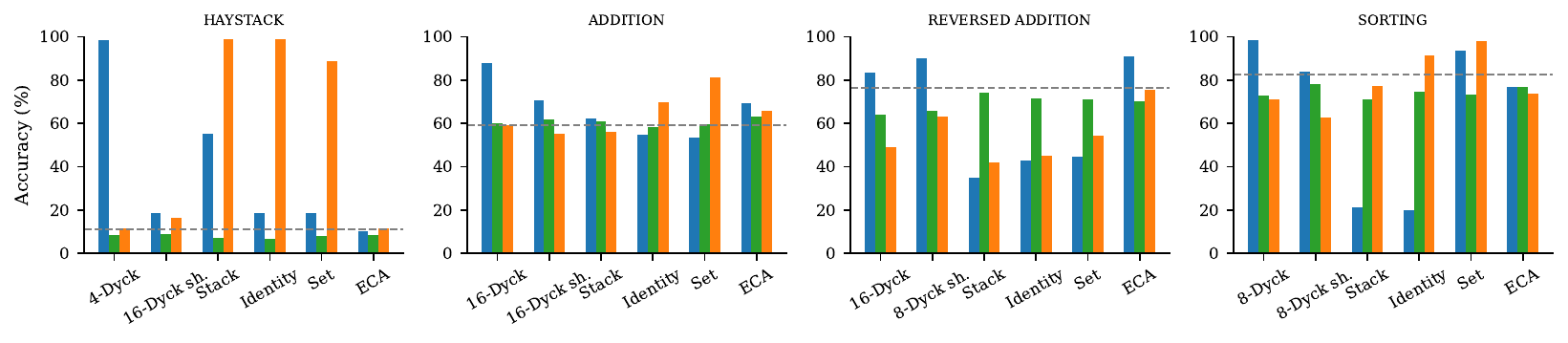}
    \end{center}
    \vspace{-16pt}
    \caption{
    Results of the selective transfer of procedurally-pretrained weights. Each color is a different type of transfer:
    \blue{\( \mathbf{\mathcal{T}_{\text{full}}} \)} 
    , \green{\( \mathbf{\mathcal{T}_{\text{mlp}}} \)} 
    , \orange{\( \mathbf{\mathcal{T}_{\text{attn}}} \)}. 
    The horizontal dotted line is the baseline performance with a random initialisation.
    See Appendix~\ref{app:additional-results} for full results including the variance across seeds.}
    \label{component-transfers}
    \vspace{-7pt}
\end{figure}

\textbf{Results.}~
Figure~\ref{component-transfers} shows that selective transfer often outperforms full
transfer, showing that \textbf{the useful inductive biases reside in specific parts of the model}.
\( \mathcal{T}_{\text{attn}}\)
frequently yields the best performance,
and in some cases substantially so.
With \textsc{Identity} pretraining for \textsc{Haystack},
\( \mathcal{T}_{\text{attn}}\)
improves by 80 percentage points over \( \mathcal{T}_{\text{full}}\)
(18.8\%\,$\rightarrow$\,99.0\%).
This indicates that the attention layers encode a useful inductive bias
for algorithmic reasoning,
while MLP weights encode mechanisms specific to the pretraining 
that do not transfer.
Similar observations are made for \textsc{Stack} and \textsc{Set} pretraining for \textsc{Haystack}.
In contrast, on \textsc{Reversed addition},
\( \mathcal{T}_{\text{mlp}} \) and \( \mathcal{T}_{\text{full}} \) outperform  \( \mathcal{T}_{\text{attn}} \).

\vspace{-2pt}
These results generally show that
the inductive biases from procedural pretraining reside
in different parts of the model, and that \textbf{they can be transferred independently},
with the attention layers being the most consistently transferable carrier.

\subsection{Pretraining Creates Soft Inductive Biases in Precisely Structured Weights}
\label{sec:structure}

Transformers are sensitive to the initialization: simply adjusting the magnitude of random weights has a big impact~\cite{huang2020improving}.
We aim to distinguish
such trivial effects from the genuine learning of 
transferrable soft inductive biases.

\textbf{Setup.}~
We introduce two types of perturbations to pretrained weights
and examine the resulting performance drop after fine-tuning.
(1)~Additive Gaussian noise checks whether inductive biases are encoded in precise weight structures; 
(2)~Random per-layer weight shuffling checks for benefits merely from the weight magnitudes. Shuffling destroys any precise structure but preserves the distribution of weight values.
We apply the perturbations on the best transfer configurations from \S\ref{sec:partial_transfer}
and report a relative improvement score
($1.0$ corresponds to unperturbed pretrained weights, $0.0$ to a random initialisation,
\( \mathcal{T}_{\text{rand}} \)).

\input{figPertubation}

\section{Combining Multiple Pretraining Tasks}
\label{sec:modular}

Since pretraining tasks
act on different layers,
can we combine components and benefits from several pretrained models?

\textbf{Setup.}~
We define \(\mathcal{T}_{\text{pre}}^1\) and \(\mathcal{T}_{\text{pre}}^2\) as two models pretrained on two types of procedural data, respectively. 
We then transfer specific components from the two models (e.g.\ the attention layers
\(\mathbf{A}_{\text{pre}}\)
from
\(\mathcal{T}_{\text{pre}}^1\)
and the MLP layers
\(\mathbf{F}_{\text{pre}}\)
from
\(\mathcal{T}_{\text{pre}}^2\)) to initialise a combined model \( \mathcal{T}_{\text{comb}} \). 
The combined model is then fine-tuned on various downstream tasks.

\input{figModularTransfer}

\textbf{Results.}~
In Table~\ref{tab:modular-transfer}, we show an example with \textsc{Set} and \textsc{ECA}, where we combine the attention layers from \textsc{Set} and MLPs from \textsc{ECA}. 
Compared to other configurations that only rely on one procedural task, which fail at least on one downstream task, the combined model (last row) consistently performs reasonably well on all four tasks.
This suggests that \textbf{the useful structures can be modularly composed} into a single ``initialisation'' that facilitates fine-tuning on multiple tasks.

%% file: figPertubation.tex
\noindent
\begin{minipage}[t]{0.3\linewidth}
    \vspace{-5pt} 
    \includegraphics[width=\linewidth]{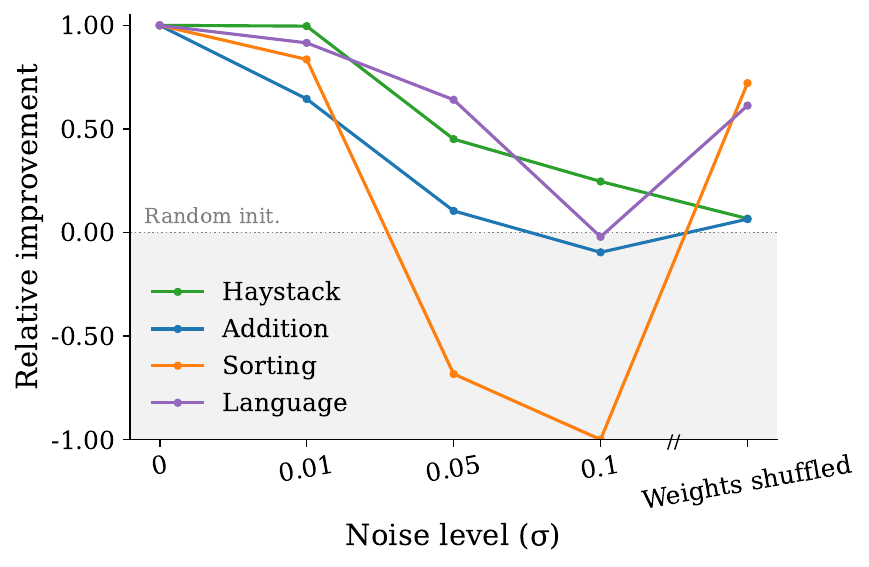}
\end{minipage}%
\hfill
\begin{minipage}[t]{0.67\linewidth}
    \vspace{-7pt}  
    \captionof{figure}{Relative downstream improvement using
    pretrained weights with perturbations (noise or shuffling). See Appendix~\ref{app:additional-results} for full results and details.}
    \label{fig:perturbation-figure}
    \vspace{0.5em}
    \textbf{Results.}~
    Gradually increasing noise consistently degrades performance, indicating that precise structure is crucial. \textsc{Haystack} and \textsc{Addition} are highly sensitive to shuffling, suggesting that \textbf{pretraining encodes information beyond weight magnitudes}. In contrast, \textsc{Language modelling} and \textsc{Sorting} are more robust, implying that \textbf{some cases partially benefit from adjusted magnitudes}.

\end{minipage}

%% file: figModularTransfer.tex
\begin{center}
\begin{minipage}[t]{0.5\linewidth}
    \vspace{-10pt}  
    \begin{small}
    \begin{sc}
    \begin{tabular}{lcccc}
    \toprule
    \textnormal{Pretraining configuration} & Haystack & Addition & Reversed addition & Sorting \\
    \midrule
    Set ~~\textnormal{(Full transfer)}       & $18.9_{\pm 26.6}$  & $53.4_{\pm 0.1}$   & $44.6_{\pm 5.1}$  & $93.5_{\pm 1.6}$ \\
    Set ~~\textnormal{(Attention only)}          & $88.9_{\pm 27.1}$  & $\mathbf{81.1}_{\pm 12.2}$   & $54.4_{\pm 10.4}$  & $98.1_{\pm 2.8}$ \\
    \midrule
    ECA ~~\textnormal{(Full transfer)}       & $10.5_{\pm 0.5}$   & $69.6_{\pm 7.9}$   & $\mathbf{91.0}_{\pm 16.1}$  & $76.9_{\pm 1.4}$ \\
    ECA ~~\textnormal{(MLPs only)}            & $8.71_{\pm 1.0}$   & $63.1_{\pm 14.4}$  & $70.5_{\pm 31.6}$  & $77.1_{\pm 8.1}$ \\
    \midrule
    Set \textnormal{(Attention)} \;+\; ECA \textnormal{(MLPs)} & $\mathbf{94.4}_{\pm 2.5}$  & $\underline{80.3}_{\pm 13.9}$   & $\underline{82.9}_{\pm 16.9}$  & $\mathbf{99.4}_{\pm 0.2}$ \\
    \bottomrule
    \end{tabular}
    \end{sc}
    \end{small}
\end{minipage}%
\hfill
\begin{minipage}[t]{0.33\linewidth}
    \vspace{-10pt}  
    \captionof{table}{Comparison of a modularly composed model with models pretrained on individual procedural data types. Combining \textsc{Set} attention and \textsc{ECA} MLP layers performs strongly on all four tasks.}
    \label{tab:modular-transfer}
\end{minipage}
\end{center}

%% file: discussion.tex
\section{Discussion and Open Questions}
\label{sec:discussion}

We showed that pretraining transformers on well-chosen procedural tasks
creates useful structure in different parts of the architecture.
We identified capabilities that significantly improve with 
specific forms of procedural data.
We also verified that the improvements cannot be explained with trivial effects
such as a better weight magnitude. 
These results open exciting questions and possibilities to take full advantage of procedural data.

\vspace{-3pt}
\textbf{Why are specific forms of procedural data helpful?}
We miss a first-principle explanation 
why specific forms of data help specific capabilities.
E.g.\ why does \textsc{Haystack} benefit from \textsc{$k$-Dyck} but not the \textsc{shuffle} variant?
How is pretraining with formal languages different from ECAs?
Our positive and negative examples could support a comparative analysis.

\vspace{-4pt}
\textbf{Combining pretraining tasks.}
It is not clear which specific forms of procedural data could help training a generalist LLM.
Data mixture optimization~\cite{fan2023doge,xie2023doremi,ye2024data}
could help balance multiple procedural rules.
This procedural data could be merged with standard pretraining data, or instead be used as a ``pre-pretraining'' curriculum.

\vspace{-4pt}
\textbf{Closed-form initialisation.}
Expanding simple procedural rules into millions of training examples
seems computationally wasteful.
Can we characterise the resulting structure in pretrained models to
directly instantiate it in initial weights?

\vspace{-4pt}
\textbf{Knowledge vs.\ reasoning.}
LLMs' difficulties to reason robustly may be rooted in entangled representations of knowledge and reasoning
\cite{han2025general}.
Procedural data could teach reasoning 
independently from
specific semantic information.

%% file: relatedWork.tex
\section{Related Work}
\label{app:related-work}

\paragraph{What is learned by pretraining language models.}
The quantity~\cite{kaplan2020scaling} and quality
\cite{longpre2024pretrainer}
of pretraining data are empirically critical for the performance of large language models.
But recent results also question the value of the data,
showing that some benefits of pretraining are attributable to the optimization objective more than the actual data.
\citet{balestriero2024perception} 
compared models trained for text classification from random initialization
with fine-tuning from a pretrained checkpoint. They found that pretraining provides little benefit for tasks that do not involve text generation.
\citet{krishna2023downstream} 
showed success in re-using the same data for pretraining and fine-tuning,
showing also that the pretraining objective matters more than the data being used.
The same conclusion follows from results of pretraining on synthetic data devoid of semantic meaning,
e.g.\ for machine translation~\cite{he2023synthetic}, 
computer vision~\cite{baradad2021learning}, 
visual navigation~\cite{wang2022visual}, 
and reinforcement learning~\cite{baradad2022procedural}. 
This paper examines such purely synthetic pretraining to understand the exact capabilities that can be obtained from procedurally-generated data.

\paragraph{What matters in pretraining data.}
The selection of data to pretrain frontier models
mostly relies on experimentation%
~\cite{longpre2024pretrainer}.
However, several key distributional and structural properties of the data
have also been identified, such as data repetition to foster generalization~\cite{charton2024emergent}
and burstiness to enable in-context learning~\cite{chan2022data}.
Computer code is empirically very effective as pretraining data for LLMs,
as it improves their abilities for compositional generalization and
math-related tasks~\cite{aryabumi2024code,petty2024does}.
This presumably results from the abundant compositional and recursive patterns in computer code,
but a better understanding of the mechanisms at play
is lacking to reap the full benefits of structure in pretraining data.
In this paper, we replicate the positive effects of structured pretraining data in controlled settings,
and study how such data imparts useful inductive biases to the model.


\paragraph{Pretraining on procedural data.%
}
Most attempts to train language models with synthetic data
follow a linguistic perspective,
using formal languages to imitate properties of natural language
\cite{chiang2022transferability,goodale2025meta,mccoy2023modeling,papadimitriou2023injecting,ri2022pretraining}.
Recent work considers increasingly simpler forms of synthetic data such as input/outputs of simple algorithms
\cite{lindemann2024sip,wu2022insights}.
In these papers, specific forms of synthetic pretraining data prove helpful to subsequent fine-tuning on natural language tasks. 
\citet{hu2025between} provide strong empirical benefits,
showing that data generated from formal languages is more valuable token-per-token than natural language
for training a 1B-parameter language model.
\citet{zhang2024intelligence} pretrain on traces of cellular automata and show marginal but consistent improvements on simple reasoning tasks.
Our study complements this line of work by examining more closely the pretrained models
on diagnostic tasks, rather than evaluating their general handling of natural language.
We identify specific capabilities imparted by specific types of procedural tasks,
and locate useful structure in different parts of the architecture.
We also investigate methods to combine the benefits from multiple complementary tasks.






\paragraph{Procedural data in vision and RL.}
Vision transformers (ViTs)
have been trained on synthetic data
of increasingly simple nature~\cite{baradad2021learning}.
\citet{nakamura2024scaling} 
pretrained ViTs on a single fractal image with augmentations
that remarkably match the performance of ImageNet-pretrained models after fine-tuning.
This indicates that 
structural properties of the data
matter more than its semantic contents.
Similar results exist in {reinforcement learning} with models pretrained
on data generated from random Markov chains
\cite{wang2023pretraining} 
and noise-based images
\cite{baradad2022procedural}. 

\paragraph{Partial transfer from pretrained transformers.}
\citet{zhang2023instilling} 
and
\cite{xu2023initializing}
showed that copying
subsets of pretrained weights
could transfer specific capabilities. 
\citet{abnar2020transferring} 
used knowledge distillation to transfer the inductive biases of one architecture into another.
The ``mimetic initialization'' of self-attention
\cite{trockman2023mimetic}
is a procedure  handcrafted to imitate the locality bias of pretrained models.
We also evaluate the partial transfer of pretrained weights,
which reveals that different pretraining tasks create useful structure in different parts of the architecture.


\paragraph{Pretraining as an inductive bias.}
Pretraining transformers on synthetic data has been used to
mimic the inductive biases of Bayesian inference 
\cite{muller2021transformers} 
and Solomonoff Induction
\cite{grau2024learning}. 
\citet{goodale2025meta} showed that well-chosen formal languages can teach complex mechanisms (e.g.\ counters) to a sequence model.
Pretraining can generally be seen as a \emph{soft} inductive bias for subsequent fine-tuning.
But there is a large gap in our understanding of its effects compared to those of
\emph{hard} inductive biases
of neural architectures~\cite{teney2024neural,teney2025we}.
\citet{han2025general} argue that the difficulties of LLMs to reason robustly
is due to their entangled representation of knowledge and reasoning.
Much remains to be understood about how both are learned from the same data~\cite{ruis2024procedural}.
Our results suggest that procedural data could be one way to acquire
reasoning mechanisms independently from specific pieces of knowledge.

%% file: appendix.tex
\section{Experimental Details}
\label{app:experiment-details}

\subsection{Procedural Data Generation}
\label{app:pretraining-corpora-details}

\textsc{$k$-Dyck.}
We generate sequences consisting of correctly formed, nested parentheses using $k$ distinct bracket pairs, where each bracket is treated as an individual token. Thus the vocabulary size is $2k$. For our experiments, we evaluate models with $k \in {4, 8, 16}$, and fix all training sequences to a length of 128 tokens. Each sequence is constructed incrementally using a stack-based approach that enforces syntactic validity. At each step, the generator samples either an opening or a closing bracket based on a predefined probability ($p_{\text{open}} = 0.49$) following \citet{papadimitriou2023injecting}. We ensure that every opened bracket is eventually closed. Specifically, if the number of remaining tokens equals the number of currently open brackets, the generator switches to exclusively emitting the appropriate closing brackets to guarantee a balanced sequence. 

\vspace{4pt}
\textsc{$k$-Dyck shuffle.} Shares the same vocabulary of opening and closing parentheses as \textsc{$k$-Dyck}, but relaxes the structural constraint of well-nestedness. Every open token has a corresponding close but they do not need to be properly nested.
We use the implementation provided by \citet{hu2025between}. Like \textsc{$k$-Dyck} we always use a sequence length of 128 tokens and evaluate $k \in {4, 8, 16}$. Our opening token probability is ($p_{\text{open}} = 0.50$). As stated by \citet{hu2025between}, the final sequence may be invalid due to truncation, but we also did not see any negative consequences of this.

\vspace{4pt}
\textsc{Stack.} Consists of sequences that simulate stack-based operations, where the first part of the input encodes a sequence of \texttt{push} and \texttt{pop} operations and the second part represents the resulting stack contents. Tokens are pushed onto a stack with a 75\% probability in the first two-thrids of the sequence and popped with 75\% probability in the later one third portion. Each push inserts a unique token, and pops remove the top of the stack. We ensure only a single occurance of a token can be present on the stack at any point in time. The input sequence is followed by a separator token and the remaining stack contents (in top-to-bottom order). The model is trained to autoregressively predict the stack tokens after the separator. We train using curriculum learning: starting with input sequences of length 4, we increase the length by 2 once the model achieves 99\% accuracy on the current sequence length. The curriculum progresses until reaching a maximum sequence length of 20. This gradually exposes the model to increasingly complex stack manipulations, allowing it to build algorithmic competence over time.  The vocabulary size is set to 103, comprising 100 pushable tokens, a dedicated \texttt{pop} token, a separator token, and a padding token.

\vspace{4pt}
\textsc{Identity.} The setup mirrors that of the \textsc{Stack}, also employing curriculum learning. Each input sequence consists of randomly sampled tokens, which are concatenated with a separator token. The target output is an exact copy of the input sequence. The model is trained to autoregressively reproduce the input tokens following the separator, requiring it to replicate the input structure. The vocabulary is set to 102 tokens, comprising 100 valid input elements, along with dedicated tokens for padding and separation.

\vspace{4pt}
\textsc{Set.} Also uses curriculum learning to progressively increase input length. Each input sequence consists of randomly sampled tokens and is followed by a separator token. The target output is a de-duplicated version of the input sequence, preserving the original order of first occurrence for each token. This requires the model to remember which elements have already appeared while autoregressively generating the output. The vocabulary size is 102, with 100 tokens representing valid input elements and two special tokens for the separator and padding.

\vspace{4pt}
\textsc{ECA Rule 110.} We adopt the training setup and codebase released by \citet{zhang2024intelligence}, where data is procedurally generated from Elementary Cellular Automata (ECA) using Rule 110, a Class IV rule known for its complex, Turing-complete behavior. To enable next-token prediction over binary state sequences, their approach modifies the standard GPT-2 architecture by replacing the token embedding layer with a linear projection that maps binary vectors directly into the model’s embedding space. Similarly, the output softmax is replaced by a linear projection back to binary space. This design allows the model to process raw binary sequences and remain deterministic, aligning with the deterministic nature of ECA dynamics. For transfer into our downstream tasks, we compute and extract the average of the learned input embeddings over the ECA pretraining data. These averaged embeddings are then used to initialise the embedding layers of our target transformer models, enabling effective transfer of structure without relying on a pretrained embedding layer.

\clearpage

\subsection{Procedural Pretraining}
\label{app:pretrain_details}

We list in Table~\ref{tab:pretraining-config} the hyperparameters for each type of procedural data other than \textsc{ECA}.

\begin{table}[h!]
\vspace{-3pt}
\begin{center}
\begin{small}
\begin{sc}
\begin{tabular}{lccccccc}
\toprule
\textnormal{Task} & Seq. length & Batch size & Learning rate & Warmup steps & Vocab. size & Early stopping & Max steps \\
\midrule
\textsc{$k$-Dyck}         & 128     & 256 & $1 \times 10^{-4}$ & 100{,}000 & $2 \times k$     & —                      & 1{,}000{,}000 \\
\textsc{K Dyck Shuffle} & 128     & 256 & $1 \times 10^{-4}$ & 100{,}000 & $2 \times k$     & —                      & 1{,}000{,}000 \\
\textsc{Stack}          & 4–20    & 256 & $5 \times 10^{-4}$ & 1{,}000   & 103              & 100 validation checks  & 1{,}000{,}000 \\
\textsc{Set}            & 2–20    & 256 & $5 \times 10^{-4}$ & 1{,}000   & 102              & 100 validation checks  & 1{,}000{,}000 \\
\textsc{Identity}       & 4–20    & 256 & $5 \times 10^{-4}$ & 1{,}000   & 102              & 100 validation checks  & 1{,}000{,}000 \\
\bottomrule
\end{tabular}
\end{sc}
\end{small}
\end{center}
\vspace{-7pt}
\caption{
Pretraining hyperparameters for each procedural task. All use AdamW and a weight decay of $0.01$.
}
\label{tab:pretraining-config}
\vspace{-2pt}
\end{table}

\textsc{ECA Rule 110.}~ 
We adopt the training configuration from \citet{zhang2024intelligence}. Models are pretrained for next-token prediction using data generated from ECA rule 110. In each epoch, a fresh dataset is generated from a new random initial state. This setup effectively simulates infinite data. Training proceeds for up to 10,000 epochs with early stopping based on validation loss. We use a batch size of 64 (60 time steps, 100 spatial dimensions), the Adam optimiser with a learning rate of $2 \times 10^{-6}$, weight decay of 0.01, and gradient clipping (norm $\leq 1.0$). A learning rate warm-up over the first 10\% of training steps is followed by cosine annealing.
 
\subsection{Description of Downstream Diagnostic Tasks}
\label{app:downstream_tasks}
\textsc{Haystack.} We adopt the task as implemented by \citet{zhong2024algorithmic} that is publicly available.
This task assesses a model’s ability to perform retrieval over long sequences. Each input consists of a series of key-value pairs in the format \([m_1, c_1, m_2, c_2, \ldots, m_k, c_k, m_u]\), where each \(m_i\) is a distinct marker, and each \(c_i\) is the corresponding value. The sequence concludes with a query marker \(m_u\), and the model is required to locate its earlier occurrence and return the associated value \(c_u\). For all experiments we set $k = 30$. Accuracy is computed on \(c_u\).

\textsc{Addition.}
This task requires the model to learn the structure of arithmetic addition presented in forward (non-reversed) notation. This is seen as a harder task than the reversed addition for transformers, as the least significant digits, which are critical to determining carry operations, appear later in the sequence. This forces the model propagate carry information backward through the input, which is misaligned with the auto retrogressive training procedure. Each input sequence takes the form \texttt{a+b=} where \(a\) and \(b\) are randomly sampled integers with a digit length $n$  and the output is their sum. Inputs and outputs are tokenised at the digit level, with symbol tokens (\texttt{+}, \texttt{=}) assigned unique indices. Models are trained to predict only the result digits, with cross-entropy loss computed exclusively on those output positions. For all experiments we set $n$ = 5. Accuracy is computed at the token level on the result digits.

\textsc{Reversed Addition.}  We again use the implementation by \citet{zhong2024algorithmic}. This task evaluates a model’s ability to perform multi-step arithmetic by adding two length $n$ integers represented as sequences of digits. To simplify the positional dependencies, both the inputs and the output are reversed: for instance, the sum \(ab + cd = efg\) is encoded with input \texttt{b a d c} and output \texttt{g f e}. The model must predict the digit-wise sum in left-to-right order, reflecting carry propagation across digit positions. We set $n = 10$ and accuracy is computed at the token level.

\textsc{Multiplication.}  
We evaluate the model’s ability to perform multi-digit multiplication. Each input sequence represents an equation of the form \(a \times b =\), where \(a\) and \(b\) are randomly sampled $n$-digit integers. The model is required to output the digits of their product. Inputs and outputs are tokenised at the digit level, with the multiplication operator \texttt{\texttimes} and the equals sign \texttt{=} assigned special token IDs. For all experiments, we set \(n = 5\). Loss and accuracy are computed only over the output portion of the sequence corresponding to the product digits.

\textsc{Sorting.}
This task evaluates a model's ability to perform algorithmic reasoning by sorting a sequence of integers. The input consists of a list of $n$ integers sampled uniformly from the range \([0, P-1]\), where \(P\) is the size of the symbol vocabulary. We set $n = 10$ and $P = 100$. The model receives the input sequence followed by a separator token and is trained to output the sorted version of the input immediately after the separator. For example, given an input sequence \texttt{6 3 5} and separator \texttt{|}, the expected output is \texttt{3 5 6}. The model is trained autoregressivley and is evaluated only on the tokens following the separator, where we calculate accuracy at the token level.

\textsc{Language Modelling.}
This task uses the \textit{TinyStories} dataset~\citep{eldan2023tinystories}, a collection of short, synthetically generated English-language narratives. Each story consists of simple sentences intended for early readers. We frame this as a last-token prediction task. The model is given the first 63 tokens of a 64-token sequence sampled from a passage and is trained to predict the 64\textsuperscript{th} token. To simplify the vocabulary, we restrict the model to the top 2{,}000 most frequent tokens from the dataset. Accuracy is computed based on whether the predicted token matches the correct token at position 64, testing the model's ability to use contextual information for natural language completion. We adapt code from the \texttt{Pluto} repository~\footnote{\url{https://github.com/tanaydesai/pluto}} for this task.

\subsection{Downstream Training}
\label{sec:downstream-training-details}
\textsc{Haystack}, \textsc{Forward addition}, \textsc{Reversed addition}, and \textsc{Sorting}.~
We trained models for $10^4$ steps with a batch size of 1,000. The training data is generated dynamically. We used the AdamW optimizer with a learning rate of $10^{-3}$ and weight decay of $10^{-3}$. We always use an architecture consisting of 2 layers, 4 attention heads, and 16-dimensional embeddings. We report mean and standard deviation over 10 seeds. The main body of the paper reports only mean accuracy for clarity; the full statistics including standard deviations are presented in Appendix~\ref{app:additional-results}.

\textsc{Multiplication}.~
These experiments employed a larger model with 4 layers, 8 attention heads, and 512-dimensional embeddings. Thus, we use a smaller training batch size (64 vs. 1,000), resulting in approximately 156k update steps compared to 10k steps for the afforementioned reasoning tasks, despite using the same number of training examples. We optimize with AdamW using a learning rate of $10^{-3}$, weight decay of $10^{-3}$, and 500 warmup steps. We run this over 3 seeds, and report standard deviations in Appendix~\ref{app:additional-results}.

\textsc{Language modelling.}~
We used a larger architecture with 2 layers, 4 attention heads, and 64-dimensional embeddings. Models are trained for 1 epoch on 1.2 million \textit{TinyStories} sequences of length 64 using a batch size of 64 and vocabulary size of 2,000. The learning rate is set to $2 \times 10^{-3}$ with 10\% linear warmup steps and cosine decay. We also run this over 10 seeds, and report standard deviations in Appendix~\ref{app:additional-results}.

\clearpage
\section{Additional Results}
\label{app:additional-results}

\begin{table}[H]
\vskip 0.15in
\begin{center}
\begin{small}
\begin{sc}
\begin{tabular}{lcccccc}
\toprule
\textnormal{Pretraining task} & Haystack & Addition & Reversed addition & Multiplication & Sorting & Language modelling \\
\midrule
Rand init.                & $11.3 \pm 0.4$ & $59.1 \pm 7.0$ & $76.4 \pm 23.2$ & $42.7 \pm 5.3$ & $82.7 \pm 11.6$& $40.2 \pm 0.2$ \\
\midrule
4-Dyck              & $98.3 \pm 1.1$ & $52.7 \pm 0.3$ & $35.7 \pm 2.5$  & $46.7 \pm 4.6$ & $56.3 \pm 19.2$& $41.6 \pm 0.2$ \\
8-Dyck              & $93.6 \pm 1.3$ & $53.4 \pm 0.3$ & $48.9 \pm 4.9$  & $44.5 \pm 0.9$ & $98.7 \pm 0.3$ & $41.2 \pm 0.2$ \\
16-Dyck             & $96.9 \pm 1.0$ & $87.8 \pm 4.2$ & $83.5 \pm 0.6$  & $39.4 \pm 3.3$ & $95.5 \pm 1.0$ & $41.0 \pm 0.5$ \\
\midrule
4-Dyck shuffle       & $7.3 \pm 0.6$  & $54.5 \pm 0.2$ & $87.8 \pm 12.9$ & $41.8 \pm 3.7$ & $61.0 \pm 1.4$ & $41.0 \pm 0.3$ \\
8-Dyck shuffle       & $9.6 \pm 0.3$  & $67.7 \pm 0.8$ & $90.1 \pm 5.9$  & $37.4 \pm 0.1$ & $84.1 \pm 5.7$ & $41.2 \pm 0.3$ \\
16-Dyck shuffle      & $18.6 \pm 26.3$& $70.8 \pm 5.5$ & $87.0 \pm 12.8$ & $44.0 \pm 0.1$ & $71.1 \pm 5.4$ & $41.2 \pm 0.3$ \\
\midrule
Stack               & $55.2 \pm 39.3$ & $62.3 \pm 5.3$ & $34.9 \pm 0.2$  & $46.6 \pm 2.0$ & $21.3 \pm 0.6$ & $38.4 \pm 0.3$ \\
\midrule
Identity            & $18.8 \pm 14.3$ & $54.7 \pm 0.2$ & $42.7 \pm 0.9$  & $46.6 \pm 2.7$ & $19.9 \pm 0.5$ & $37.9 \pm 0.2$ \\
\midrule
Set                 & $18.9 \pm 26.6$ & $53.4 \pm 0.1$ & $44.6 \pm 5.1$  & $43.5 \pm 8.4$ & $93.5 \pm 1.6$ & $40.05 \pm 0.5$ \\
\midrule
ECA                 & $10.5 \pm 0.5$ & $69.6 \pm 7.9$ & $91.1 \pm 16.1$ & ---            & $76.9 \pm 1.4$ & --- \\
\bottomrule
\end{tabular}
\end{sc}
\end{small}
\end{center}
\vskip -0.1in
\caption{Full results across all pretraining tasks and downstream tasks. Each cell reports the mean accuracy $\pm$ standard deviation over 10 random seeds, except for \textsc{Multiplication} and \textsc{Language modelling}, which is over 3 seeds. A subset of these results is visualised in Figure~\ref{first-figure}.}

\label{tab:full_res}
\end{table}

\begin{table}[H]
\label{tab:NIAH-component-transfer}
\vskip 0.15in
\begin{center}
\begin{small}
\begin{sc}
\begin{tabular}{lccc}
\toprule
\textnormal{Pretraining task} & Full transfer & MLP only & Attention only \\
\midrule
4-Dyck            & $98.3 \pm 1.1$   & $8.7 \pm 0.5$   & $11.6 \pm 0.5$ \\
16-Dyck shuffle    & $18.6 \pm 26.3$  & $8.9 \pm 0.9$   & $16.5 \pm 10.6$ \\
Stack             & $55.2 \pm 39.3$  & $7.1 \pm 0.6$   & $98.9 \pm 0.8$ \\
Identity          & $18.8 \pm 14.3$  & $7.0 \pm 0.9$   & $99.0 \pm 1.7$ \\
Set               & $18.9 \pm 26.6$  & $8.3 \pm 0.7$   & $88.9 \pm 27.1$ \\
ECA               & $10.5 \pm 0.5$   & $8.7 \pm 1.0$   & $11.6 \pm 1.0$ \\
\bottomrule
\end{tabular}
\end{sc}
\end{small}
\end{center}
\vskip -0.1in
\caption{
\textsc{Haystack} task accuracy (mean $\pm$ standard deviation over 10 seeds) for models initialised with weights from different pretraining tasks. We report results for full model transfer, MLP-only transfer, and attention-only transfer.
}
\end{table}

\begin{table}[H]
\label{tab:addition-component-transfer}
\vskip 0.15in
\begin{center}
\begin{small}
\begin{sc}
\begin{tabular}{lccc}
\toprule
\textnormal{Pretraining task} & Full transfer & MLP only & Attention only \\\midrule
16-Dyck           & $87.8 \pm 4.2$   & $60.0 \pm 6.6$   & $59.2 \pm 10.4$ \\
16-Dyck shuffle    & $70.8 \pm 5.5$   & $61.7 \pm 6.9$   & $55.3 \pm 4.9$ \\
Stack             & $62.3 \pm 5.3$   & $61.1 \pm 9.4$   & $56.2 \pm 5.0$ \\
Identity          & $54.7 \pm 0.2$   & $58.3 \pm 7.2$   & $69.7 \pm 13.1$ \\
Set               & $53.4 \pm 0.1$   & $59.6 \pm 6.4$   & $81.1 \pm 12.2$ \\
ECA               & $69.6 \pm 7.9$   & $63.1 \pm 14.4$  & $65.8 \pm 12.8$ \\
\bottomrule
\end{tabular}
\end{sc}
\end{small}
\end{center}
\vskip -0.1in
\caption{
\textsc{Addition} task accuracy (mean $\pm$ standard deviation over 10 seeds) for models initialised with weights from different pretraining tasks. We report results for full model transfer, MLP-only transfer, and attention-only transfer.
}
\end{table}

\begin{table}[H]
\label{tab:reversedaddition-component-transfer}
\vskip 0.15in
\begin{center}
\begin{small}
\begin{sc}
\begin{tabular}{lccc}
\toprule
\textnormal{Pretraining task} & Full transfer & MLP only & Attention only \\\midrule
16-Dyck            & $83.5 \pm 0.6$   & $64.0 \pm 26.4$  & $49.1 \pm 20.3$ \\
8-Dyck shuffle        & $90.1 \pm 5.9$   & $65.8 \pm 24.8$  & $63.3 \pm 18.1$ \\
Stack             & $34.9 \pm 0.2$   & $74.4 \pm 24.7$  & $42.1 \pm 8.1$ \\
Identity          & $42.7 \pm 0.9$   & $71.7 \pm 29.2$  & $45.2 \pm 3.7$ \\
Set               & $44.6 \pm 5.1$   & $71.2 \pm 23.7$  & $54.4 \pm 10.4$ \\
ECA               & $91.1 \pm 16.1$  & $70.5 \pm 31.6$  & $75.5 \pm 27.2$ \\
\bottomrule
\end{tabular}
\end{sc}
\end{small}
\end{center}
\vskip -0.1in
\caption{
\textsc{Reversed addition} task accuracy (mean $\pm$ standard deviation over 10 seeds) for models initialised with weights from different pretraining tasks. We report results for full model transfer, MLP-only transfer, and attention-only transfer.
}
\end{table}

\begin{table}[H]
\label{tab:sorting-component-transfer}
\vskip 0.15in
\begin{center}
\begin{small}
\begin{sc}
\begin{tabular}{lccc}
\toprule
\textnormal{Pretraining task} & Full transfer & MLP only & Attention only \\\midrule
8-Dyck         & 98.7$\pm$0.3  & 72.8$\pm$3.1  & 71.4$\pm$5.7 \\
8-Dyck shuffle  & 84.1$\pm$5.7  & 78.2$\pm$8.6  & 62.9$\pm$6.7 \\
Stack          & 21.3$\pm$0.6  & 71.0$\pm$2.2  & 77.5$\pm$12.2 \\
Identity       & 19.9$\pm$0.5  & 74.5$\pm$8.1  & 91.3$\pm$10.1 \\
Set            & 93.5$\pm$1.6  & 73.5$\pm$1.5  & 98.1$\pm$2.8 \\
ECA            & $76.9 \pm 1.4$  & 77.1$\pm$8.1 & 73.9$\pm$3.2 \\
\bottomrule
\end{tabular}
\end{sc}
\end{small}
\end{center}
\vskip -0.1in
\caption{
\textsc{Sorting} task accuracy (mean $\pm$ standard deviation over 10 seeds) for models initialised with weights from different pretraining tasks. We report results for full model transfer, MLP-only transfer, and attention-only transfer.
}
\end{table}

\begin{table}[h]
\centering

\label{tab:perturbation-table}
\vspace{0.5em}
\begin{small}
\begin{tabular}{lccccc}
\toprule
\textnormal{Perturbation} & \textsc{Haystack} & \textsc{Addition} & \textsc{Reversed addition} & \textsc{Sorting} & \textsc{Language modelling} \\
\midrule
Pretrained   & $98.9 \pm 0.8$ & $87.8 \pm 4.2$ & $90.1 \pm 5.9$ & $98.7 \pm 0.3$ & $41.6 \pm 0.2$ \\
Shuffled     & $17.2 \pm 12.7$ & $61.0 \pm 9.1$ & $82.9 \pm 23.5$ & $94.2 \pm 4.2$ & $41.1 \pm 0.4$ \\
0.01 noise   & $98.6 \pm 1.7$ & $77.6 \pm 20.1$ & $74.0 \pm 21.0$ & $96.0 \pm 7.6$ & $41.5 \pm 0.5$ \\
0.05 noise   & $50.8 \pm 30.5$ & $62.1 \pm 13.3$ & $91.0 \pm 15.7$ & $71.9 \pm 26.1$ & $41.1 \pm 0.3$ \\
0.10 noise   & $32.9 \pm 6.1$ & $56.4 \pm 7.4$ & $83.6 \pm 21.5$ & $37.9 \pm 5.8$ & $40.2 \pm 0.1$ \\
Random init  & $11.3 \pm 0.4$ & $59.1 \pm 7.0$ & $76.4 \pm 23.2$ & $82.7 \pm 11.6$ & $40.2 \pm 0.2$ \\
\bottomrule
\end{tabular}
\caption{
Mean accuracy ($\pm$ standard deviation over 10 seeds) across five downstream tasks under different perturbation conditions. Pretrained models were selected based on best individual performance per task: \textsc{Stack} (attention-only) for \textsc{Haystack}, \textsc{16-Dyck} for \textsc{Addition}, \textsc{8-Dyck shuffle} for \textsc{Reversed addition}, \textsc{8-Dyck} for \textsc{Sorting}, and \textsc{4-Dyck} for \textsc{Language Modelling}.
}

\end{small}
\end{table}